\documentclass[lettersize,journal]{IEEEtran}
\usepackage{amsmath,amsfonts}
\usepackage{algorithmic}
\usepackage{algorithm}
\usepackage{array}
\usepackage[caption=false,font=normalsize,labelfont=sf,textfont=sf]{subfig}
\usepackage{textcomp}
\usepackage{stfloats}
\usepackage{url}
\usepackage{verbatim}
\usepackage{graphicx}
\usepackage{cite}
\usepackage{fontawesome}
\usepackage{mathrsfs}
\usepackage{multirow}
\begin{document}

\title{Learning Accurate and Enriched Features for Stereo Image Super-Resolution}

\author{Hu Gao and Depeng Dang
\thanks{Hu Gao and Depeng Dang are with the School of
Artificial Intelligence, Beijing Normal University,
Beijing 100000, China (e-mail: gao\_h@mail.bnu.edu.cn, ddepeng@bnu.edu.cn).}}



\maketitle

\begin{abstract}
Stereo image super-resolution (stereoSR) aims to enhance the quality of super-resolution results by incorporating  complementary information from an alternative view. Although current methods have shown significant advancements, they typically operate on representations at full resolution to preserve  spatial details, facing challenges in accurately capturing contextual information. Simultaneously, they utilize all feature similarities to cross-fuse information from the two views, potentially disregarding the impact of irrelevant information. To overcome this problem, we propose a mixed-scale selective fusion network (MSSFNet) to preserve precise spatial details and incorporate abundant contextual information, and adaptively select and fuse most accurate features from two views to enhance the promotion of high-quality stereoSR. 
Specifically, we develop a mixed-scale block (MSB) that obtains contextually enriched feature representations across multiple spatial scales while preserving precise spatial details. Furthermore, to dynamically retain the most essential cross-view information, we design a selective fusion attention module (SFAM) that searches and transfers the most accurate features from another view. To learn an enriched set of local and non-local features, we introduce a fast fourier convolution block (FFCB) to explicitly integrate frequency domain knowledge.  Extensive experiments show that MSSFNet achieves significant improvements over state-of-the-art approaches on both quantitative and qualitative evaluations. The code and the pre-trained models will be released at~\url{https://github.com/Tombs98/MSSFNet}.
\end{abstract}


\section{Introduction}
\IEEEPARstart{S}{tereo} image super resolution (stereo SR)  strives to improve  the resolution  from a pair of low-resolution (LR) left and right images. It can be defined as follows:
\begin{equation}
\label{equ:define}
SR_L, SR_R = F(LR_L, LR_R)
\end{equation}
where $LR_L$ and $LR_R$ refer to the left and right LR images, respectively, while $SR_L$ and $SR_R$ denote their corresponding SR images. The stereo SR function is denoted by $F(\cdot)$. It relies not only on the intra-view information within the $LR_L, LR_R$  but also on the cross-view information between them. Consequently, relying on established single image SR (SISR) methods, such as~\cite{tcsvt10440324,edsrlim2017enhanced, rdnzhang2018residual, rcanzhang2018image, swinirliang2021swinir, Singlea9786841, Singleli2022real}, for independent obtain of the $SR_L, SR_R$ images  suffers inferior performance due to the absence of cross-view information. 

\begin{figure}[t] 
	\centering
	\includegraphics[width=1\linewidth]{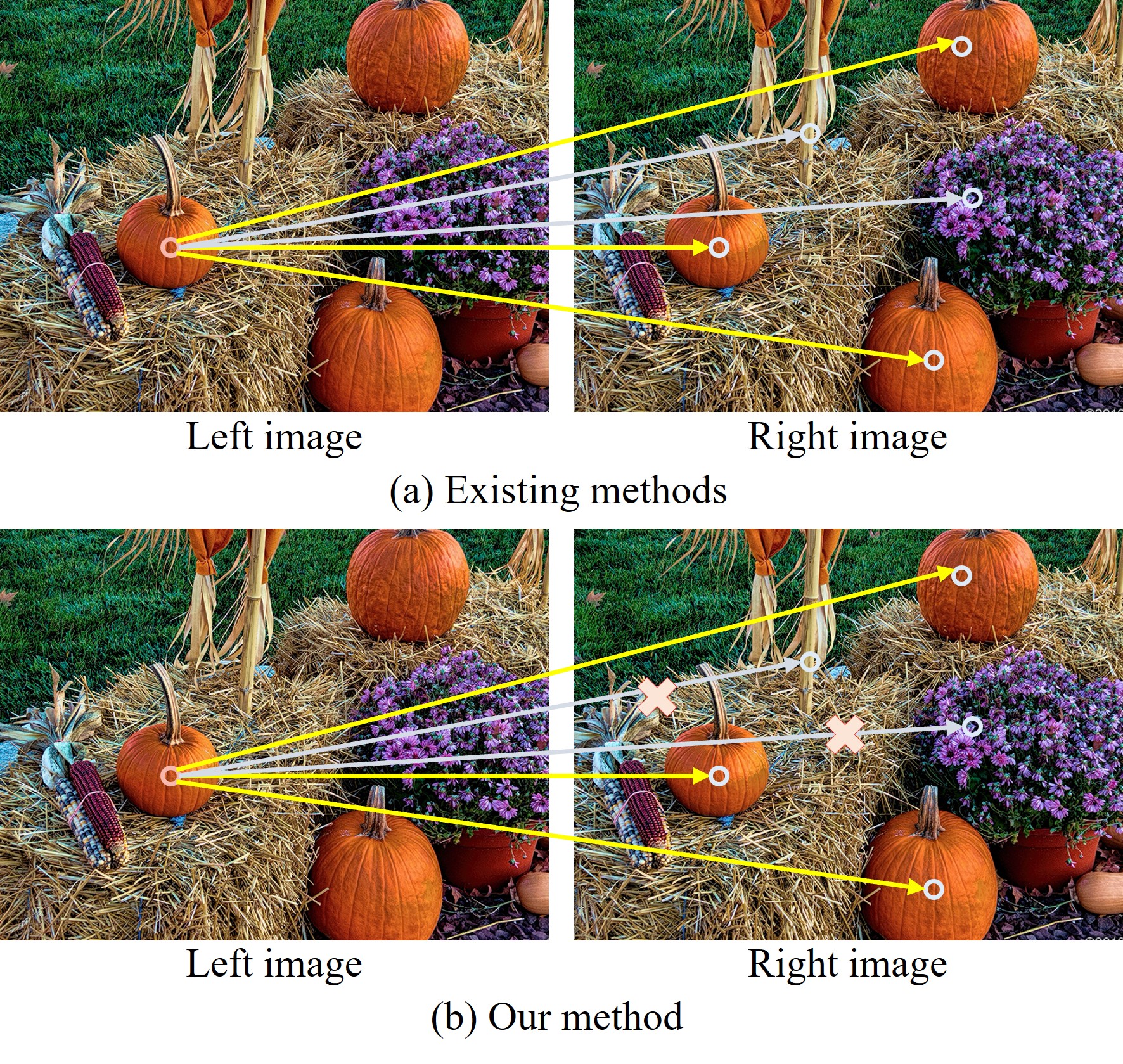}
	\caption{Comparison between our method and existing methods. Existing methods utilize all attention relations based on another viewpoint to aggregate features. Our method searches and transfers the most accurate features from another view to reduce the distraction of irrelevant information. }
	\label{fig:att}
\end{figure}

To enhance the performance of stereoSR, numerous efforts~\cite{tcsvt9940619,sirformeryang2022sir, nafssrchu2022nafssr, swinfsrChen_2023_CVPR, steformer10016671, imssrnet9253563, stereoma2021perception, bassrnet9382858, mspfinet10182355,flickr1024wang2019learning,ipassrwang2021symmetric} focus on refining the network structure to efficiently extract intra-information and design cross-attention feature fusion modules to aggregate additional cross-information from an alternative viewpoint. PASSRnet~\cite{flickr1024wang2019learning} first attempt to introduce a parallax-attention stereo super resolution network, which employs a global receptive field along the epipolar line to effectively handle a diverse set of stereo images exhibiting substantial disparities.~\cite{ipassrwang2021symmetric} introduces a symmetric bi-directional parallax attention module (biPAM) and an integrated occlusion handling scheme for the symmetrical super-resolution of both the left and right images.~\cite{cvcss9465749} devise a cross-view block and integrated a cascaded spatial perception module to proficiently capture more effective features from both global and local viewpoints.~\cite{nafssrchu2022nafssr} tackles the issue of model complexity in existing methods by introducing NAFSSR, which seamlessly blends the simplicity and effectiveness of NAFNet~\cite{nafblockchen2022simple} with the distinctive attributes of stereo super-resolution tasks through the use of stereo cross-attention modules (SCAM). To effectively capture global information,~\cite{swinfsrChen_2023_CVPR} explicitly integrating frequency domain knowledge into the Residual Swin Transformer blocks (RSTBs) and refining the biPAM, eliminating occlusion handling while redesigning the attention mechanism.~\cite{steformer10016671} introduces a cross-attentive feature extraction module and a cross-to-intra information integration module to effectively capture dependable stereo correspondence and seamlessly integrate cross-view information for stereo image super-resolution.

While the aforementioned methods have achieved significant advancements, they still exhibit certain limitations.  
On the one hand, they typically operate on representations at full resolution (single-scale)~\cite{steformer10016671, scvscaai2023joint, cpassrnet9318556, cvcss9465749}, excelling in generating images with spatially accurate details. Nonetheless, their effectiveness in encoding contextual information is hindered by their limited receptive field. In fact, it is worth noting that rich contextual information has proven highly effective in enhancing super-resolution performance~\cite{Zamir2022MIRNetv2}.
On the other hand, as shown in Figure~\ref{fig:att}, most cross-attention modules~\cite{nafssrchu2022nafssr,swinfsrChen_2023_CVPR,ipassrwang2021symmetric, scvscaai2023joint, sirformeryang2022sir} typically utilize all attention relations based on another viewpoint to aggregate features. The underlying issue here is that not all features from another viewpoint can provide beneficial complementary information (see gray line), making the feature interaction and aggregation process vulnerable to implicit noises. This naturally results in the consideration of corresponding redundant or irrelevant representations when modeling global feature dependencies. 

\begin{figure*}[t] 
	\centering
	\includegraphics[width=1\linewidth]{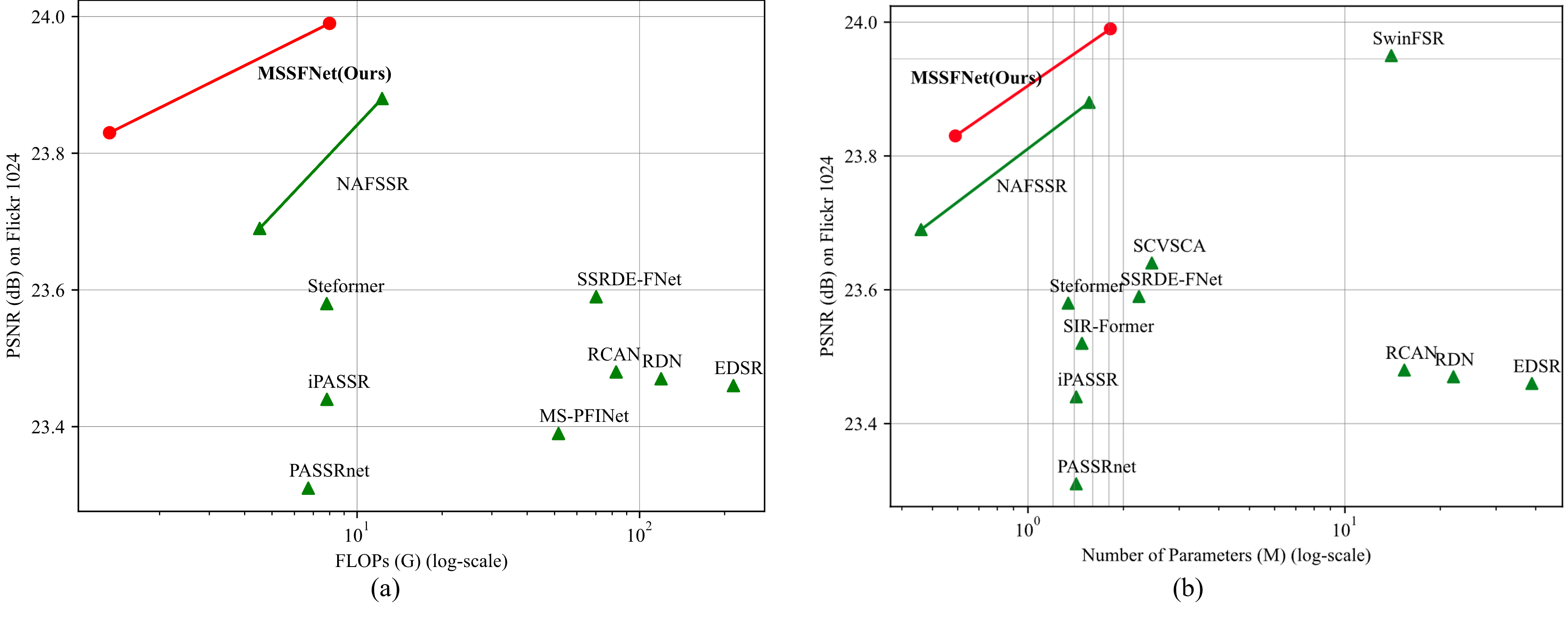}
	\caption{ Computational cost vs. PSNR between our MSSFNet and other state-of-the-art algorithms for 4× stereo SR on Flickr 1024~\cite{flickr1024wang2019learning}. (a) Our MSSFNet achieve the SOTA performance with fewer FLOPs. (b) The total number of parameters vs. PSNR. Our MSSFNet achieve the best performance
 with up to 87\% of parameter reduction.}
	\label{fig:param}
\end{figure*}

Based on the information presented, a natural question that comes to mind is whether it is feasible to design a network that effectively learns rich  intra-view information while selectively fuses the most accurate cross-view information? To this end, we develop a mixed-scale selective fusion network for stereoSR, named as MSSFNet.
Specially, we first design a mixed-scale block (MSB) to  effectively captures enriched intra-view features. It initial operate on the original high-resolution features, thus maintains precise spatial details. And then, it incorporates two multi-scale depth-wise convolution paths to encode multi-scale context features. This enables us to obtain more precise spatial details and contextually enriched feature representations. 
Secondly, we propose a selective fusion attention module (SFAM) to adaptively aggregate  cross-view information. Unlike the existing methods~\cite{nafssrchu2022nafssr,swinfsrChen_2023_CVPR}, we start by generating global feature descriptors that merge information from both the left and right views. Following this, we utilize these descriptors to modify the feature maps of each view, extracting the most useful "selected" features. Subsequently, a fusion attention module is applied to aggregate these "selected" features, efficiently removing any irrelevant information that might cause interference. 
Thirdly, we introduce a fast fourier convolution block (FFCB)   to integrate both local and global information. The FFCB consists of two interconnected branches: a local branch that performs  on  input feature channels, and a global branch that operates in the spectral domain. These two branches independently capture complementary information and subsequently engage in fusion exchange to derive the final result.
As shown in Figure~\ref{fig:param}, our  MSSFNet model achieves state-of-the-art performance while also maintaining a fewer computational cost in comparison to existing methods.

 The main contributions of this work are:
\begin{enumerate}
\item  We present a  mixed-scale selective fusion network for stereoSR, named as MSSFNet, which effectively  learns rich intra-view information while selectively fusing the most accurate cross-view information. Extensive experiments are conducted to demonstrate the effectiveness and efficiency of our proposed MSSFNet.

\item  We propose a novel mixed-scale block (MSB) designed to acquire contextually enriched feature representations across multiple spatial scales while preserving precise spatial details.

\item We design a selective fusion attention module (SFAM) to adaptively maintain the most accurate information exchange between intra-view and cross-view features.

\item We introduce a fast fourier convolution block (FFCB) to explicitly integrate frequency domain knowledge that provide collaborative refinement for the MSB.

\end {enumerate}

\section{Related Works}
\subsection{Single Image Super-resolution}
The challenge of Single Image Super-Resolution (SISR) has been a focal point of research for decades~\cite{tcsvt10440324,10098834,Singlea9786841,Singleli2022real, Singlepark2021dynamic, Singleyang2020learning}, aiming to generate high-resolution images solely from intra-view information derived from their low-resolution counterparts. The pioneering work of SRCNN~\cite{srcnndong2014learning} marked the initiation of applying deep learning to SISR, employing a three-layer convolutional neural network for the task. As the pursuit of enhanced representation capabilities continues, more intricate models have been devised. VDSR~\cite{vdsrkim2016accurate} and EDSR~\cite{edsrlim2017enhanced} not only increase the depth and width of the model but also incorporate skip connections for residual information learning, effectively preventing issues like gradient collapse. Addressing the contextual aspects of SISR, CBAM~\cite{cbamwoo2018cbam} employs channel and spatial attention blocks, proving highly effective in extracting contextual relations. 
JWSGN~\cite{9786841} advocates for the inclusion of frequency information in image super-resolution by utilizing wavelet transform (WT) to separate different frequency components and employing a multi-branch network for their recovery.
RWSR-EDL~\cite{li2022real} addresses the neglect of the relationship between L1 and perceptual minimization in deep learning-based image super-resolution  to enhance feature diversity in cooperative learning.
In  order to capture global information, Transformers~\cite{swinirliang2021swinir, transformervaswani2017attention, Swinfirzhang2022swinfir} have been applied to SISR, showcasing commendable performance. SwinIR~\cite{swinirliang2021swinir}, in particular, introduces an image restoration method based on the Swin Transformer~\cite{swintransformerliu2021swin}, achieving state-of-the-art performance in SISR.  While SISR has demonstrated promising performance, it faces challenges in stereoSR where the absence of cross-view information leads to inferior results.

\subsection{Stereo Image Super-resolution}
Stereo image super-resolution (stereoSR) aims to enhance high-resolution details in both the left and right views of stereo image pairs by incorporating cross-view information~\cite{stereowang2022ntire, stereowang2023ntire,tcsvt9940619}. 
While StereoSR~\cite{stereosrjeon2018enhancing} learns a parallax prior by jointly training cascaded sub-networks, it faces challenges in scenes with significant disparity variations due to fixed shift intervals.
To address this limitation, PASSRnet~\cite{flickr1024wang2019learning} introduces a parallax attention module for stereo correspondence with a global receptive field along the epipolar line. 
iPASSR~\cite{ipassrwang2021symmetric} incorporates a symmetric bi-directional parallax attention module (biPAM) and an inline occlusion handling scheme to effectively utilize symmetry cues in the super-resolution of both left and right images.
CVCnet~\cite{cvcss9465749} seamlessly incorporates cross-view spatial features from both global and local perspectives.
SSRDE-FNet~\cite{ssrdefnetdai2021feedback} efficiently addresses stereoSRand disparity estimation concurrently within a unified framework, fostering mutual enhancement between the two tasks.
NAFSSR~\cite{nafssrchu2022nafssr} achieves remarkable results by integrating simple cross-view attention modules (SCAMs) between consecutive NAFBlocks~\cite{nafblockchen2022simple}.
SWCVIN~\cite{swcvinhe2023strong} introduces a robust strong-weak cross-view interaction mechanism to surpass the maximum SR performance achieved by utilizing strong and weak cross-view interactions independently.
Transformer-based models~\cite{sirformeryang2022sir, steformer10016671, swinfsrChen_2023_CVPR} are now gaining traction in stereoSR for their ability to capture long-range dependencies. 
SIR-Former~\cite{sirformeryang2022sir} pioneers the use of transformers in stereo image super-resolution, employing a cross-attention module to learn epipolar line relationships and a transformer-based fusion module for accurate cross-view feature integration. SwinFSR~\cite{swinfsrChen_2023_CVPR} extends the StereoSR method, incorporating frequency domain knowledge through fast Fourier convolution~\cite{fouriersuvorov2022resolution}.
Furthermore, Steformer~\cite{steformer10016671} leverages Transformer's self-attention to capture both cross-view and intra-view information in stereo images, ensuring reliable stereo correspondence and effective cross-view integration. 

While the aforementioned models demonstrate substantial performance improvements, they either overlook multi-scale information or neglect the interference from irrelevant information during cross-view fusion. Therefore, in this paper,  we propose a mixed-scale selective fusion network (MSSFNet) to obtains contextually enriched feature representations across multiple spatial scales while preserving precise spatial details, and adaptively select and fuse most accurate features from two views to enhance the promotion of high-quality stereoSR.

\section{Method}
\begin{figure*}[htb] 
	\centering
	\includegraphics[width=1\textwidth]{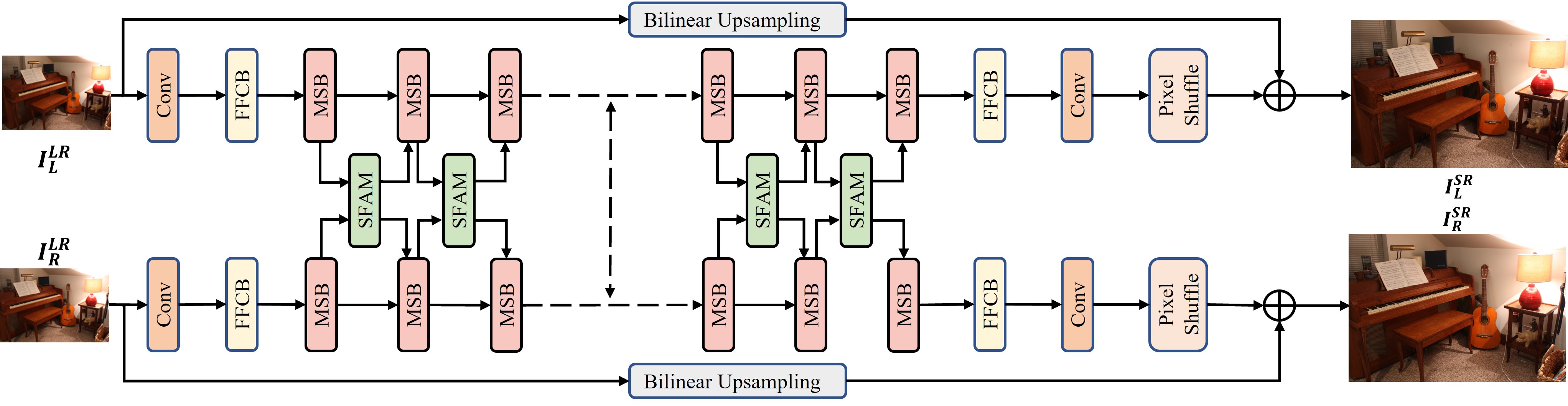}
	\caption{The overall architecture of MSSFNet with two key conponent: (1) mixed-scale block (MSB) (illustrated  in Figure.~\ref{fig:conponent}(a) ), fast fourier convolution block (FFCB) (depicted in Figure.~\ref{fig:conponent}(c))   and selective fusion attention module (SFAM) (depicted in Figure.~\ref{fig:conponent}(d)) }
	\label{fig:network}
\end{figure*}

Our primary objective is to explore a network that effectively learns rich  intra-view information while selectively fusing the most accurate cross-view information. To achieve this, we present a mixed-scale selective fusion network (MSSFNet) shown in Figure.~\ref{fig:network}. MSSFNet takes a low-resolution stereo image pair as input and enhances the resolution of both the left and right view images. It employs two weight-sharing branches constructed with mixed-scale block (MSB) to obtain more precise spatial details and contextually enriched intra-view feature representations. Furthermore, selective fusion attention modules (SFAMs) are incorporated to identify and transfer the most accurate cross-view features. Additionally, fast fourier convolution block (FFCB) are employed to integrate both local and global information.

\noindent\textbf{Overall Pipeline.} 
Given a pair of low resolution stereo images $\mathbf{I^{LR}_L} \in \mathbb R^{H \times W \times 3}$ (left view) and $\mathbf{I^{LR}_R} \in \mathbb R^{H \times W \times 3}$ (right view), MSSFNet first applies a $3 \times 3$ convolutional layer to extract shallow feature maps $\mathbf{F_{L}^S} \in \mathbb R^{H \times W \times C}$,  $\mathbf{F_{R}^S} \in \mathbb R^{H \times W \times C}$ ($H, W, C$ are the feature map height, width, and channel number, respectively).  In the network backbone, we stack $N$ MSBs to achieve deep intra-view feature extraction. Additionally, we integrate SFAM after each MSB to facilitate the interaction with cross-view information. At the early and final stages of the model learning, we introduce FFCB to explicitly integrate frequency domain knowledge that provide collaborative refinement for MSB. Following the aforementioned process, we acquire deep features denoted as $\mathbf{F_{L}^D}$ and $\mathbf{F_{R}^D}$, encompassing both intra-view and cross-view information.  Furthermore, we apply a $3 \times 3$ convolution layer followed by a pixel shuffle layer to upsample the deep feature by a scale factor of $s$, and generate $\mathbf{R_L}\in  R^{H \times W \times 3}$, $\mathbf{R_R}\in  R^{H \times W \times 3}$. 
Noted, to alleviate the burden of feature extraction, the  $\mathbf{R_L}\in  R^{H \times W \times 3}$, $\mathbf{R_R}\in  R^{H \times W \times 3}$ is the difference between the bilinearly upsampled low-resolution image and the high-resolution ground truth. Thus, the $\mathbf{I^{SR}_L} = \mathbf{R_L} + Up(\mathbf{I^{LR}_L})$, $\mathbf{I^{SR}_R} = \mathbf{R_R} + Up(\mathbf{I^{LR}_R})$ are the high-resolution images of the left and right views, respectively.

\subsection{Mixed-scale block (MSB)}
\begin{figure*}[htb] 
	\centering
	\includegraphics[width=1\textwidth]{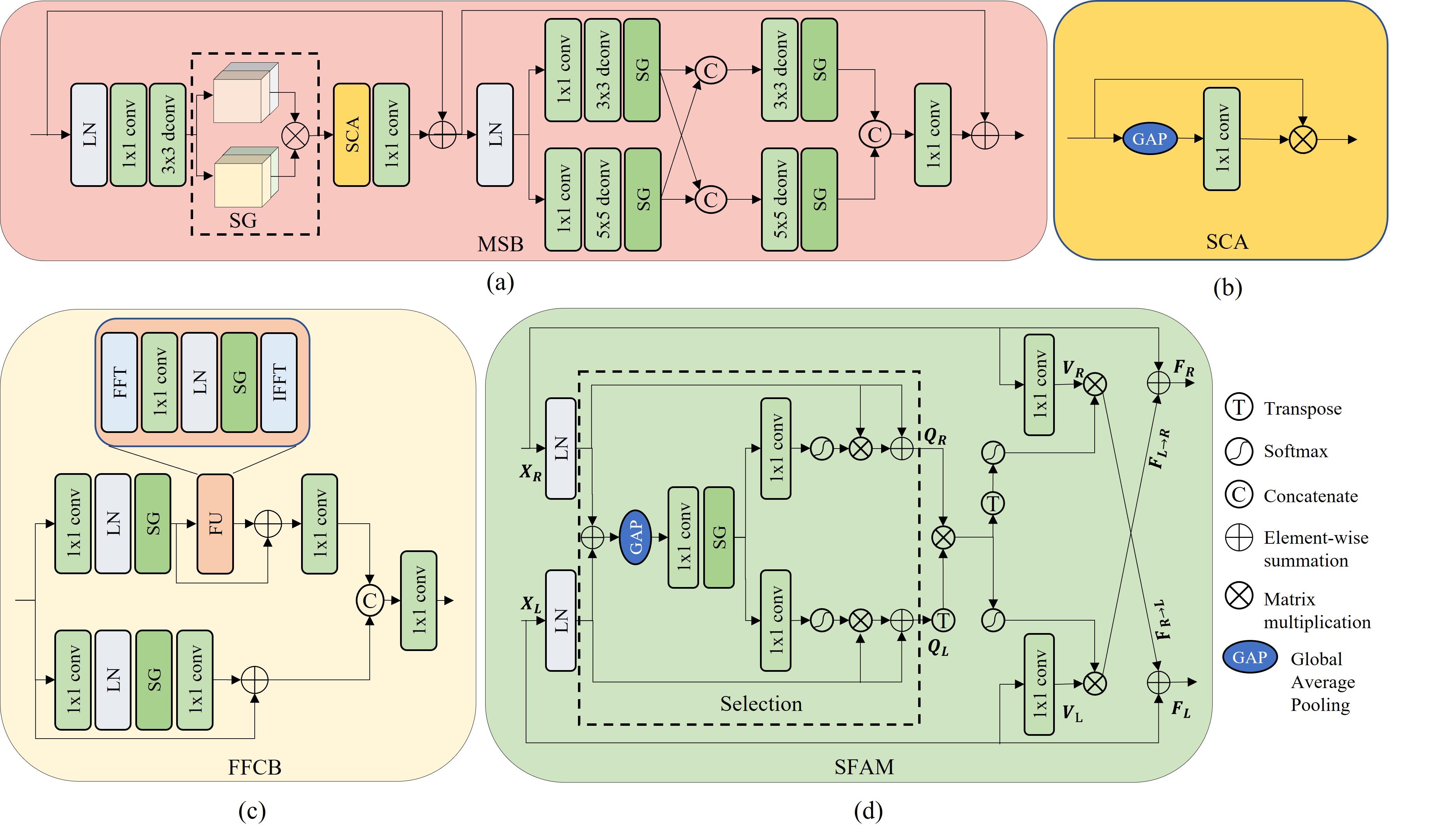}
	\caption{(a) Mixed-scale block (MSB). (b) Simplified Channel Attention (SCA). (c) Fast fourier convolution block (FFCB). (d) Selective fusion attention module (SFAM). }
	\label{fig:conponent}
\end{figure*}

Earlier works~\cite{nafssrchu2022nafssr, ipassrwang2021symmetric, cvcss9465749, scvscaai2023joint} typically operate on representations at full resolution (single-scale), resulting in images with more spatially accurate details. Nevertheless, these networks may be less effective in encoding contextual information. Indeed, the effectiveness of rich multi-scale representations has been fully demonstrated~\cite{Zamir2020MIRNet, Zamir2022MIRNetv2} in improving super-resolution performance. In this context, we  design a mixed-scale block (MSB) to simultaneously preserve precise spatial details and acquire contextually enriched feature representations across multiple spatial scales.
As shown in Figure.~\ref{fig:conponent}(a), the MSB initially operates on the original high-resolution features, thereby preserving precise spatial details. Subsequently, it integrates two multi-scale depth-wise convolution paths to encode multi-scale contextual features. Specifically, given an input tensor at the $(l-1)_{th}$ block $X_{l-1} \in \mathbb R^{H \times W \times C}$, we initially process it at the original resolution through Layer Normalization (LN), Convolution, Simple Gate (SG), and Simplified Channel Attention (SCA) to obtain spatially detailed features $X^{s}_{l-1}$ as follows:
\begin{equation}
\begin{aligned}
	\label{equ:0msb}
	X^{'}_{l-1} &= SCA(SG(f_{3 \times 3}^{dwc} (f_{1 \times 1}^c(LN(X_{l-1}))))
\\
	X^{s}_{l-1} &= f_{1 \times 1}^c(X^{'}_{l-1})+X_{l-1}
\\
    SCA(X_{f3}) &= X_{f3} \otimes f_{1 \times 1}^c( GAP(X_{f3}))
\\
    SG(X_{f0}) &= X_{f1} \otimes X_{f2} 
\end{aligned}
\end{equation}
where  $f_{1 \times 1}^c$ represents $1 \times 1$ convolution, $f_{3 \times 3}^{dwc}$ denotes the $3 \times 3$ depth-wise convolution, $\odot$ represents element-wise multiplication, and GAP is the global average pooling. $SG(\cdot)$ represents the simple gate, employed as a replacement for the nonlinear activation function. For a given input $X_{f0}$, SG initially splits it into two features $X_{f1}, X_{f2} \in \  R^{H \times W \times \frac{C}{2}}$  along channel dimension. Subsequently, SG calculates the $X_{f1}, X_{f2}$ using a linear gate. 
For a more intuitive representation, we illustrate $SCA(\cdot)$ in Fig.\ref{fig:conponent}(b). SCA is a simplified version of SE\cite{hu2018squeeze}, which aggregates global information and facilitates channel information interaction.

Subsequently, following layer normalization for the spatially detailed features $X^{s}_{l-1}$, we start by utilizing a 1 × 1 convolution to expand the channel dimension. Subsequently, we feed it into two parallel branches to handle features of different scales. Finally, the outputs from the two branches are fused to obtain a contextually enriched feature $X^{c}_{l-1}$.  The entire procedure can be formulated as:
\begin{equation}
\begin{aligned}
	\label{equ:1msb}
	X^{t1}_{l-1} &= SG(f_{3 \times 3}^{dwc} (f_{1 \times 1}^c (LN(X^{s}_{l-1}))))
 \\
 X^{b1}_{l-1} &= SG(f_{5 \times 5}^{dwc} (f_{1 \times 1}^c (LN(X^{s}_{l-1}))))
 \\
 X^{t2}_{l-1} &= SG(f_{3 \times 3}^{dwc}([X^{t1}_{l-1}, X^{b1}_{l-1}]))
 \\
 X^{b2}_{l-1} &= SG(f_{5 \times 5}^{dwc}([X^{b1}_{l-1}, X^{t1}_{l-1}]))
 \\
 X^{c}_{l-1} &=  f_{1 \times 1}^c([X^{t2}_{l-1}, X^{b2}_{l-1}])
\end{aligned}
\end{equation}
where $[\cdot]$ represents the channel-wise concatenation. Finally, we obtain precise spatial details and  contextually enriched feature representations $X_l$ as follows:
\begin{equation}
    \label{equ:2msb}
    X_l = X^{s}_{l-1} + X^{c}_{l-1}
\end{equation}

\subsection{Selective fusion attention module (SFAM)}
We revisit  all the cross-attention modules proposed in prior works~\cite{ipassrwang2021symmetric, nafssrchu2022nafssr, swinfsrChen_2023_CVPR}, which have become a common empirical operation in the majority of existing models. They all based on Scaled Dot Product Attention~\cite{NIPS2017_3f5ee243}, which computes the dot products of the query $Q\in \mathbb R^{H \times W \times C }$, the value $V \in \mathbb R^{H \times W \times C }$  projected by the source intra-view feature (e.g., left-view) with the key $K \in \mathbb R^{H \times W \times C }$ projected using the target intra-view feature (e.g., right-view).  Followed by applying a softmax function  to obtain the weights on the values:

\begin{equation}
	\label{equ:102}
    Attention(Q, K, V) =  softmax(\frac{QK^T}{\beta})V
\end{equation}
where $\beta$ is an optional temperature factor used to adjust the magnitude of the dot product of $Q$ and $K$ prior to the application of the softmax function defined by $\beta = \sqrt{C}$. 
It is worth noting that the vanilla attention module calculates the similarity between all features of the left and right views. However, not all features from another viewpoint can provide useful complementary information. This makes the feature interaction and aggregation process susceptible to implicit noise. Consequently, it is natural to consider the inclusion of corresponding redundant or irrelevant representations when modeling global feature dependencies. As shown in Figure~\ref{fig:att}(a), existing attention methods incorporate the object of the left and right view connected by gray lines into the attention calculation. This uncorrelated representation will introduce implicit noise into the model.  In our work, we propose a selective fusion attention module (SFAM) to adaptively maintain the most accurate information exchange between intra-view and cross-view features, thereby removing any irrelevant information during the cross-view feature interaction process (see Figure~\ref{fig:att}(b)).  As illustrated in Figure~\ref{fig:conponent}(d), our SFAM involves two processes: \textbf{Selection} and \textbf{Attention}.

\textbf{Selection.} We first execute a selection operation to searches the most relevant features from another view. Subsequently, we transfer these "selected " features to the cross-view feature fusion process. Specifically, given a pair of intra-view features $X_L, X_R \in \mathbb R^{H \times W \times C}$, we begin by applying layer normalization, and subsequently combine these left and right view features using an element-wise sum as:
\begin{equation}
	\label{equ:s1sum}
    X_S = LN(X_L) + LN(X_R)
\end{equation}

Then we apply global average pooling (GAP) across the spatial dimension of$X_ S\in \mathbb R^{H \times W \times C}$ to compute channel-wise statistics. 
Subsequently, we apply a channel-downscaling convolution layer and SG to generate a compact feature representation. We pass this compact feature through two parallel channel-upscaling convolution layers to create global feature descriptors that merge information from both the left and right views. Following this, we use these descriptors to modify the feature maps of each view, extracting the most accurate "selected" features. In this way, the entire feature select procedure is  defined as:
\begin{equation}
\begin{aligned}
\label{equ:select}
    X^{c}_S &= SG(f_{1 \times 1}^c(GAP(X_S)))
    \\
    X^{d}_L &= softmax(f_{1 \times 1}^c(X^{c}_S))
    \\
    X^{d}_R &= softmax(f_{1 \times 1}^c(X^{c}_S))
      \\
    Q_L &=  LN(X_L) + LN(X_L) \otimes X^{d}_L
    \\
    Q_R &=  LN(X_R) + LN(X_R) \otimes X^{d}_R
\end{aligned}
\end{equation}

\textbf{Attention.} Noted that stereo images are highly symmetric under epipolar constraint~\cite{ipassrwang2021symmetric}, we use $K_L = Q_R$ and  $K_R = Q_L$ to represent each intra-view features. Next, we generate the value matrix $V_L$ and $V_R$ by using a $1 \times 1$ convolution layer, respectively. Subsequently, we compute bidirectional cross-attention between the left and right views as follows:
\begin{equation}
    \begin{aligned}
     F_{L \rightarrow R} &= Attention(Q_L, K_L, V_L)
     \\
     F_{R \rightarrow L} &= Attention(Q_R, K_R , V_R)
    \end{aligned}
\end{equation}

Finally, the interacted cross-view information $F_{R \rightarrow L}$, $ F_{L \rightarrow R}$ and intra-view information $X_L$, $X_R$ are fused by element-wise addition:
\begin{equation}
    \begin{aligned}
        F_L &= X_L + \lambda_L F_{R \rightarrow L}
        \\
        F_R &= X_R + \lambda_R F_{L \rightarrow R}
    \end{aligned}
\end{equation}
where $\lambda_L$ and $\lambda_R$ are channel-wise scale parameters that are trainable and initialized with zeros to aid in stabilizing training.

\subsection{Fast fourier convolution block (FFCB)}
To create a more lightweight network architecture, our backbone network is based on CNN. However, due to the limited receptive field, CNNs may struggle to capture global information effectively. Yet, integrating both local and global information is crucial for improving stereoSR performance~\cite{9578694}. To empower our network with this capability, we introduce a fast fourier convolution block (FFCB) which exhibits the main characteristics of non-local receptive fields. This block explicitly integrates frequency domain knowledge, offering collaborative refinement for the MSB. As shown in Figure.~\ref{fig:conponent}(c), the FFCB comprises two branches: a local spatial branch that performs regular convolutions on  input feature channels at the bottom, and a global Fourier unit (FU) that operates in the spectral domain at the top. The outputs from both operations are concatenated and then undergo a convolution operation to generate the final result. Specifically, given an input tensor $F \in \mathbb R^{H \times W \times C}$, we send it into two distinct branches. In the bottom branch, a residual connection and two convolution layers are inserted to extract the local features $F_L$ in the spatial domain as follows:
\begin{equation}
\label{equ:ffcbl}
	F_L = f_{1 \times 1}^c (SG(LN(f_{1 \times 1}^c(F)))) + F
\end{equation}

In the top branch, we broaden the receptive field of the convolution to the full resolution of the input feature map. We transform the conventional spatial features into the frequency domain to extract global features using  FFT. Afterward, we perform the inverse  FFT operation to produce the final spatial domain features $X_G$ as follows:

\begin{equation}
\begin{aligned}
	\label{equ:ffcbg}
	F^{'} &= SG(LN(f_{1 \times 1}^c(F)))
 \\
 F_F & = \mathcal{F}^{-1}(SG(LN(f_{1 \times 1}^c(\mathcal{F}(F^{'}))))) + F^{'}
 \\
 F_G &=  f_{1 \times 1}^c(F_F)
\end{aligned}
\end{equation}
where $\mathcal{F}(\cdot)$ is the channel-wise  FFT operation, and $\mathcal{F}^{-1}(\cdot) $ is the inverse  FFT operation.

After obtaining the features from the two branches, we concatenate them and then employ a 1×1 convolutional layer to fuse the two features. This step reduces the number of channels by half, resulting in the final feature $F_{FFCB}$ that contains both local and non-local information.

\begin{equation}
\label{equ:ffcblg}
	F_{FFCB} = f_{1 \times 1}^c([F_L, F_G])
\end{equation}

\subsection{Training Strategies}
\noindent\textbf{Data augmentation.} In stereoSR tasks, it is a common practice to introduce random horizontal and vertical flips to increase dataset diversity. Additionally, for improved data utilization, we incorporate channel shuffling, which involves randomly rearranging the RGB channels to augment the colors. Additionally, we employ stochastic depth~\cite{huang2016deep} as a regularization.

\noindent \textbf{Loss function.}  We only use the pixel-wise L1 distance between
the super-resolved and ground truth stereo image: 
\begin{equation}
    L = \|I_L^{SR} - I_L^{HR} \|_1 + \|I_R^{SR} - I_R^{HR} \|_1
\end{equation}
where $I_L^{SR}$ and $I_R^{SR}$ represent the super-resolved left and
right images, and $I_L^{HR}$ and $I_R^{HR}$ represent their corresponding ground-truth high-resolution images.

\section{Experiments}

\subsection{Implementation Details}

\textbf{Datasets.} 
In training the proposed network, we utilize a dataset consistent with that of~\cite{ipassrwang2021symmetric, nafssrchu2022nafssr} to ensure a fair comparison with prior research. Specifically, our training dataset comprises 800 images from the Flickr1024 dataset~\cite{flickr1024wang2019learning} and 60 images from the Middlebury dataset~\cite{middleburyscharstein2014high}. Due to the significantly higher spatial resolution of Middlebury dataset images, we conduct bicubic downsampling with a scale factor of $2$ to generate High-Resolution (HR) images. For Low-Resolution (LR) image generation, we apply bicubic downsampling to HR images using scaling factors of $2\times$ and $4\times$. Subsequently, the resulting LR images are cropped into $30\times90$ patches with a stride of $20$, and their HR counterparts undergo corresponding cropping. This process yields a total of $49,020$ patches for $4\times$ Super-Resolution (SR) training and $298,143$ patches for $2\times$ SR training.
To evaluate the proposed network's performance, we use a test dataset consisting of $20$ images from the KITTI 2012 dataset~\cite{kitti2012geiger2012we}, and $20$ images from the KITTI 2015 dataset~\cite{kitti2015menze2015object}, $112$ images from the Flickr1024 dataset~\cite{flickr1024wang2019learning}, $5$ images from the Middlebury dataset~\cite{middleburyscharstein2014high}. 

\textbf{Evaluation details.}  To ensure a fair comparison with~\cite{nafssrchu2022nafssr, ipassrwang2021symmetric, flickr1024wang2019learning, swinfsrChen_2023_CVPR}, we evaluate Peak Signal-to-Noise Ratio (PSNR) and Structural Similarity (SSIM) specifically on the left views, with the left boundaries cropped ($64$ pixels). Additionally, we compute the average scores on stereo image pairs, calculated as ($\text{Left} + \text{Right}) /2$, without any boundary cropping.

\textbf{Training details.}
We construct two different sizes of MSSFNet networks by adjusting the number of channels and blocks, as detailed in Table~\ref{tab:archi}. The training of MSSFNet is carried out using the Adam optimizer~\cite{2014Adam} with parameters $\beta_1=0.9$ and $\beta_2=0.9$. A fixed batch size of $32$ is employed for $1 \times 10^5$ iterations. The training commences with a learning rate of $1 \times 10^{-3}$ and gradually decreases to $1 \times 10^{-7}$ using cosine annealing~\cite{2016SGDR}. To mitigate overfitting, stochastic depth~\cite{huang2016deep} with a 0.1 probability is incorporated for MSSFNet-S. Specifically, since the lightweight model MSSFNet-T encounters underfitting rather than overfitting, it undergoes $4 \times$ training iterations without stochastic depth.

\begin{table}[hb]
    \caption{Architecture Variants of MSSFNet.}
    \label{tab:archi}
    \centering
    \begin{tabular}{cccc}
    \hline
         Models&  $\#Channels$&  $\#Blocks$ & $\#P$ 
         \\
         \hline
         MSSFNet-T & C = 48 & N = 16 & 0.57M
         \\
         MSSFNet-S & C = 64 & N = 32 &1.80M
         \\
         \hline
    \end{tabular}
\end{table}

\subsection{Comparison with the State-of-the-Arts}

We compare our MSSFNet with existing SR methods, encompassing both  single image SR methods~\cite{edsrlim2017enhanced, rdnzhang2018residual, rcanzhang2018image, swinirliang2021swinir}  and stereo image SR methods~\cite{stereosrjeon2018enhancing, flickr1024wang2019learning, bassrnet9382858, cpassrnet9318556, imssrnet9253563, ipassrwang2021symmetric, sirformeryang2022sir, cvcss9465749,scvscaai2023joint, ssrdefnetdai2021feedback, swcvinhe2023strong, nafssrchu2022nafssr,  steformer10016671}. Note that, for a fair comparison, we retrained all of these methods using our training dataset.

\begin{table*}
\caption{Quantitative results achieved by various methods on the $KITTI2012$~\cite{kitti2012geiger2012we}, $KITTI2015$~\cite{kitti2015menze2015object},  $Middlebury$~\cite{middleburyscharstein2014high} and $Flickr1024$~\cite{flickr1024wang2019learning} datasets. The number of network parameters denoted as $\#P$. The reported values include PSNR/SSIM results for both the left images (i.e., $Left$) and a pair of stereo images (i.e., $(Left + Right) / 2$. The best results are highlighted in bold.}
\label{tb:Quantitative}
    \centering
     \resizebox{\linewidth}{!}{
    \begin{tabular}{ccccccccccc}
         \hline
         \multirow{2}{*}{Method} &  \multirow{2}{*}{Scale} &  \multirow{2}{*}{$\#P$} & \multicolumn{3}{c}{$Left$} &&\multicolumn{4}{c}{$(Left + Right) / 2$}
         \\
         \cline{4-6}
         \cline{8-11}
          & & &KITTI2012& KITTI2015&  Middlebury& &KITTI2012& KITTI2015&  Middlebury&Flicker1024
          \\
          \hline
          EDSR~\cite{edsrlim2017enhanced} & x2 & 38.6M &30.83/0.9199 &29.94/0.9231 &34.84/0.9489 &&30.96/0.9228 &30.73/0.9335 &34.95/0.9492 &28.66/0.9087
          \\
          RDN~\cite{rdnzhang2018residual} & ×2 &22.0M &30.81/0.9197 &29.91/0.9224 &34.85/0.9488 &&30.94/0.9227 &30.70/0.9330 &34.94/0.9491 &28.64/0.9084
          \\
          RCAN~\cite{rcanzhang2018image}& ×2 &15.3M &30.88/0.9202 &29.97/0.9231 &34.80/0.9482 &&31.02/0.9232 &30.77/0.9336 &34.90/0.9486 &28.63/0.9082
          \\
          SwinIR~\cite{swinirliang2021swinir} &x2 &1.32M &  30.89/0.9206 &29.98/0.9237 &34.69/0.9475&& 31.02/0.9235 &30.77/0.9341 &34.80/0.9478&28.67/0.9091
          \\
          StereoSR~\cite{stereosrjeon2018enhancing} &x2  &1.08M &29.42/0.9040 &28.53/0.9038 &33.15/0.9343 &&29.51/0.9073 &29.33/0.9168 &33.23/0.9348 &25.96/0.8599
          \\
          PASSRnet~\cite{flickr1024wang2019learning} &×2 &1.37M &30.68/0.9159 &29.81/0.9191 &34.13/0.9421 &&30.81/0.9190 &30.60/0.9300 &34.23/0.9422 &28.38/0.9038
          \\
          BASSRnet~\cite{bassrnet9382858} &x2 & 1.89M   &30.99/0.9225 &30.05/0.9256 &34.73/0.9468 &&31.03/0.9241 &30.74/0.9344 &34.74/0.9475& 28.53/0.9090
          \\
          CPASSR~\cite{cpassrnet9318556} &x2 & 5.26M &29.68/0.9079 &29.69/0.9193 &33.68/0.9433 &&29.87/0.9113 &30.39/0.9295 &33.85/0.9436& 28.12/0.9017
          \\
          IMSSRnet~\cite{imssrnet9253563} &x2 & 6.84M & 30.90/- &29.97/- &34.66/- &&30.92/- &30.66/- &34.67/- & -
          \\
          iPASSR~\cite{ipassrwang2021symmetric} &x2  &1.38M &30.97/0.9210 &30.01/0.9234 &34.41/0.9454 &&31.11/0.9240 &30.81/0.9340 &34.51/0.9454 &28.60/0.9097
          \\
         
           SSRDE-FNet~\cite{ssrdefnetdai2021feedback} & ×2 &2.10M &31.08/0.9224 &30.10/0.9245 &35.02/0.9508 &&31.23/0.9254 &30.90/0.9352 &35.09/0.9511 &28.85/0.9132
          \\
          CVCnet~\cite{cvcss9465749} & x2 & 0.97M&- &-&- & & 30.87/0.9198 & 29.93/0.9224 & 34.40/0.9450 & 28.44/0.9081
          \\
          SCVSCA~\cite{scvscaai2023joint} & x2 &2.46M & 30.98/0.9129 &30.04/0.9161 &34.96/0.9436&& 31.12/0.9162 &30.83/0.9273 &35.02/0.9434 &28.87/0.9035
          \\
          SWCVIN~\cite{swcvinhe2023strong} &x2 & 1.11M & - & - &- && 31.19/0.9250 &30.29/0.9450&34.99/0.9510 &-
          \\
           MS-PFINet~\cite{mspfinet10182355} &x2 &1.40M &- &- &- & &31.09/0.9230 &30.12/0.9250 &34.90/0.9490 &28.66/0.9130
          \\
           SIR-Former~\cite{sirformeryang2022sir}  &×2 &1.37M &31.02/0.9217 &30.11/0.9246 &34.87/0.9490 &&31.16/0.9247 &30.93/0.9355 &34.95/0.9495 &28.69/0.9103
          \\
          Steformer~\cite{steformer10016671} &x2 &1.29M & 31.16/0.9236 &30.27/0.9271 &35.15/0.9512 && 31.29/0.9263 &31.07/0.9371 &35.23/0.9511 &28.97/0.9141
          \\
           NAFSSR-T~\cite{nafssrchu2022nafssr} &x2&0.45M &31.12/0.9224 &30.19/0.9253 &34.93/0.9495 &&31.26/0.9254 &30.99/0.9355 &35.01/0.9495 &28.94/0.9128
          \\
          NAFSSR-S~\cite{nafssrchu2022nafssr} &x2 &1.54M &  31.23/0.9236 &30.28/0.9266 &35.23/0.9515&& 31.38/0.9266 &31.08/0.9367 &35.30/0.9514 &29.19/0.9160
          
          \\
          
       \hline
        \textbf{MSSFNet-T (Ours)} & x2 & 0.57M & 31.19/0.9235	&30.23/0.9263	&35.18/0.9511&&				31.34/0.9265	&31.03/0.9365	&35.24/0.9512	&28.97/0.9152

         \\
       \textbf{MSSFNet-S (Ours)} & x2 & 1.80M & \textbf{31.37/0.9262}	&\textbf{30.37/0.9287}	&\textbf{35.77/0.9555}&&	\textbf{31.53/0.9292}	&\textbf{31.16/0.9384}	&\textbf{35.82/0.9553} &\textbf{29.45/0.9212}

         \\
      
         \hline
         \hline
          EDSR~\cite{edsrlim2017enhanced} & x4 & 38.9M  &26.26/0.7954 &25.38/0.7811 &29.15/0.8383 &&26.35/0.8015 &26.04/0.8039 &29.23/0.8397 &23.46/0.7285
          \\
          RDN~\cite{rdnzhang2018residual} & ×4 &22.0M & 26.23/0.7952 &25.37/0.7813 &29.15/0.8387 &&26.32/0.8014 &26.04/0.8043 &29.27/0.8404 &23.47/0.7295
          \\
          RCAN~\cite{rcanzhang2018image}& ×4 &15.4M &26.36/0.7968 &25.53/0.7836 &29.20/0.8381 &&26.44/0.8029 &26.22/0.8068 &29.30/0.8397 &23.48/0.7286
          \\
          SwinIR~\cite{swinirliang2021swinir} &x4 &1.35M & 26.43/0.7996 &25.60/0.7868 &29.16/0.8379&&26.52/0.8058 &26.29/0.8098 &29.25/0.8385 &23.53/0.7322
          \\
          StereoSR~\cite{stereosrjeon2018enhancing} &x4  &1.42M &24.49/0.7502 &23.67/0.7273 &27.70/0.8036 &&24.53/0.7555 &24.21/0.7511 &27.64/0.8022 &21.70/0.6460
          \\
          PASSRnet~\cite{flickr1024wang2019learning} &×4 &1.42M & 26.26/0.7919 &25.41/0.7772 &28.61/0.8232 &&26.34/0.7981 &26.08/0.8002 &28.72/0.8236 &23.31/0.7195
          \\
          BSSRnet~\cite{bassrnet9382858} &×4 &1.91M & 26.45/0.8014 &25.57/0.7872 &29.12/0.8354 & &26.47/0.8049 &26.17/0.8075 &29.08/0.8362 &23.40/0.7289
          \\
          CPASSR~\cite{cpassrnet9318556} &×4 &5.26M &25.38/0.7753 &25.05/0.7707 &28.47/0.8245 & &25.50/0.7818 &25.63/0.7926 &28.55/0.8251 &23.12/0.7161
          \\
          IMSSRnet~\cite{imssrnet9253563} &×4 &6.89M &26.44/- &25.59/- &29.02/- &&26.43/- &26.20/- &29.02/- &-
          \\
          iPASSR~\cite{ipassrwang2021symmetric} &x4  &1.42M & 26.47/0.7993 &25.61/0.7850 &29.07/0.8363 &&26.56/0.8053 &26.32/0.8084 &29.16/0.8367 &23.44/0.7287
          \\
         
            SSRDE-FNet~\cite{ssrdefnetdai2021feedback} & ×4 &2.24M &26.61/0.8028 &25.74/0.7884 &29.29/0.8407 &&26.70/0.8082 &26.43/0.8118 &29.38/0.8411 &23.59/0.7352
          \\
           CVCnet~\cite{cvcss9465749} & x4 & 0.99M&- &-&- & & 26.35/0.7935 & 25.55/0.7801 & 28.65/0.8231 & 23.22/0.7192
           \\
          SCVSCA~\cite{scvscaai2023joint} & x4 &2.46M & 26.58/0.7864 &25.73/0.7736 &29.30/0.8286  &&26.68/0.7932 &26.44/0.7974 &29.40/0.8285  &23.64/0.7186
          \\
          SWCVIN~\cite{swcvinhe2023strong} &x4 & 1.12M & - & - &- && 26.68/0.8190 &25.81/0.8115 &29.37/0.8352 &-
          \\
           MS-PFINet~\cite{mspfinet10182355} &x4 &1.45M &- &- &- & &26.52/0.8000 &25.65/0.7860 &29.04/0.8340 &23.39/0.7300
          \\
           SIR-Former~\cite{sirformeryang2022sir}  &×4 &1.48M &26.53/0.7998 &25.75/0.7882 &29.23/0.8396 &&26.68/0.8077 &26.42/0.8098 &29.32/0.8407 &23.52/0.7305
          \\
          Steformer~\cite{steformer10016671} &x4 &1.34M & 26.61/0.8037 &25.74/0.7906 &29.29/0.8424 & &26.70/0.8098 &26.45/0.8134 &29.38/0.8425 &23.58/0.7376
          \\
          NAFSSR-T~\cite{nafssrchu2022nafssr} &×4 &0.46M &26.69/0.8045 &25.90/0.7930 &29.22/0.8403 &&26.79/0.8105 &26.62/0.8159 &29.32/0.8409 &23.69/0.7384
          \\
          NAFSSR-S~\cite{nafssrchu2022nafssr} &x4 &1.56M &  26.84/0.8086 &26.03/0.7978 &29.62/0.8482 &&26.93/0.8145 &26.76/0.8203 &29.72/0.8490 &23.88/0.7468
          \\
          \hline
          \textbf{MSSFNet-T (Ours)} &x4 &0.59M & 26.77/0.8063	&25.96/0.7946	&29.38/0.8426	&&	26.85/0.8119	&26.69/0.8173	&29.48/0.8433	&23.81/0.7418
          \\
         \textbf{MSSFNet-S (Ours)} &x4 &1.82M & \textbf{26.88/0.8098}	&\textbf{26.07/0.7990}&\textbf{29.67/0.8498}	&&	\textbf{26.97/0.8158}	&\textbf{26.82/0.8219}	&\textbf{29.77/0.8502}	&\textbf{23.99/0.7508}
            
          \\
        \hline
    \end{tabular}}
\end{table*}

\textbf{Quantitative results.}
As the quantitative results shown in Table.~\ref{tb:Quantitative}, MSSFNet  achieves considerable results on all datasets~\cite{flickr1024wang2019learning, kitti2012geiger2012we, kitti2015menze2015object ,middleburyscharstein2014high} and upsampling factors ($\times2, \times4$). These results serve as further validation of the effectiveness of our proposed method.
Specifically, in the case of $2 \times$ stereo super-resolution (SR), our MSSFNet-T outperforms the previous state-of-the-art model NAFSSR-T~\cite{nafssrchu2022nafssr} by 0.08 dB, 0.04 dB, 0.23 dB, and 0.03 dB on KITTI 2012~\cite{kitti2012geiger2012we}, KITTI 2015~\cite{kitti2015menze2015object}, Middlebury~\cite{middleburyscharstein2014high}, and Flickr 1024 datasets~\cite{flickr1024wang2019learning}, while MSSNet-S surpasses NAFSSR-S~\cite{nafssrchu2022nafssr} by 0.15 dB, 0.08 dB, 0.52 dB, and 0.26 dB, respectively. 
It is evident that as the network size increases, our MSSFNet demonstrates greater performance potential compared to NAFSSR.

\begin{figure*}
    \centering
    \includegraphics[width=1\linewidth]{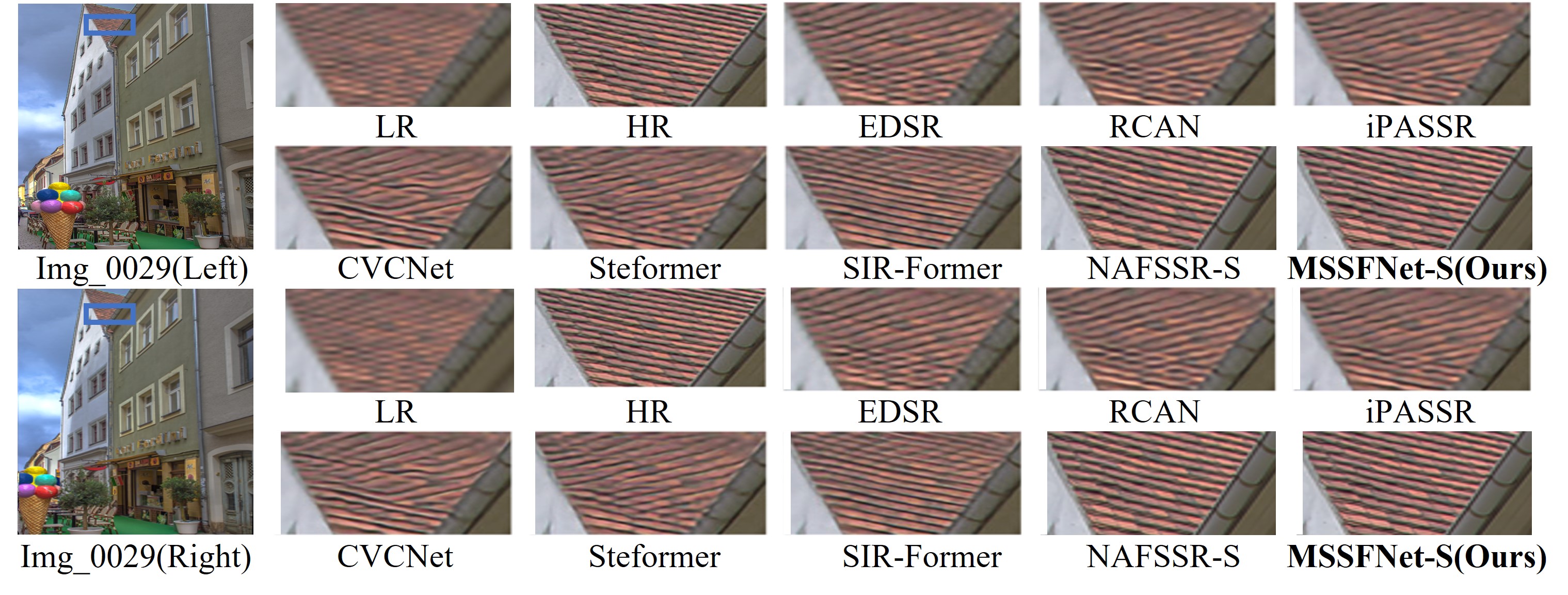}
    \caption{Visual results ($\times 2$) achieved by different methods on the Flickr1024 dataset~\cite{flickr1024wang2019learning}.}
    \label{fig:2xv}
\end{figure*}

We also presented a visualization of the trade-off results between the computational cost and PSNR on the Flickr 1024 dataset~\cite{flickr1024wang2019learning} for $4 \times$ stereo SR.  As shown in Figure~\ref{fig:param}(a), our MSSFNet achieve the best performance with fewer FLOPs.
 And Figure~\ref{fig:param}(b) clearly demonstrates that our MSSFNet outperforms SwinFSR~\cite{swinfsrChen_2023_CVPR} by achieving a state-of-the-art result while remarkably reducing the number of parameters by 87\%. Furthermore, when compared to NAFSSR~\cite{nafssrchu2022nafssr}, we achieve superior PSNR values using an equivalent number of parameters. These findings highlight the highly efficient of MSSFNet. Moreover, through the scaling up of the model size, our MSSFNet achieves even better performance, highlighting the scalability of MSSFNet.

\begin{figure*}
    \centering
    \includegraphics[width=1\linewidth]{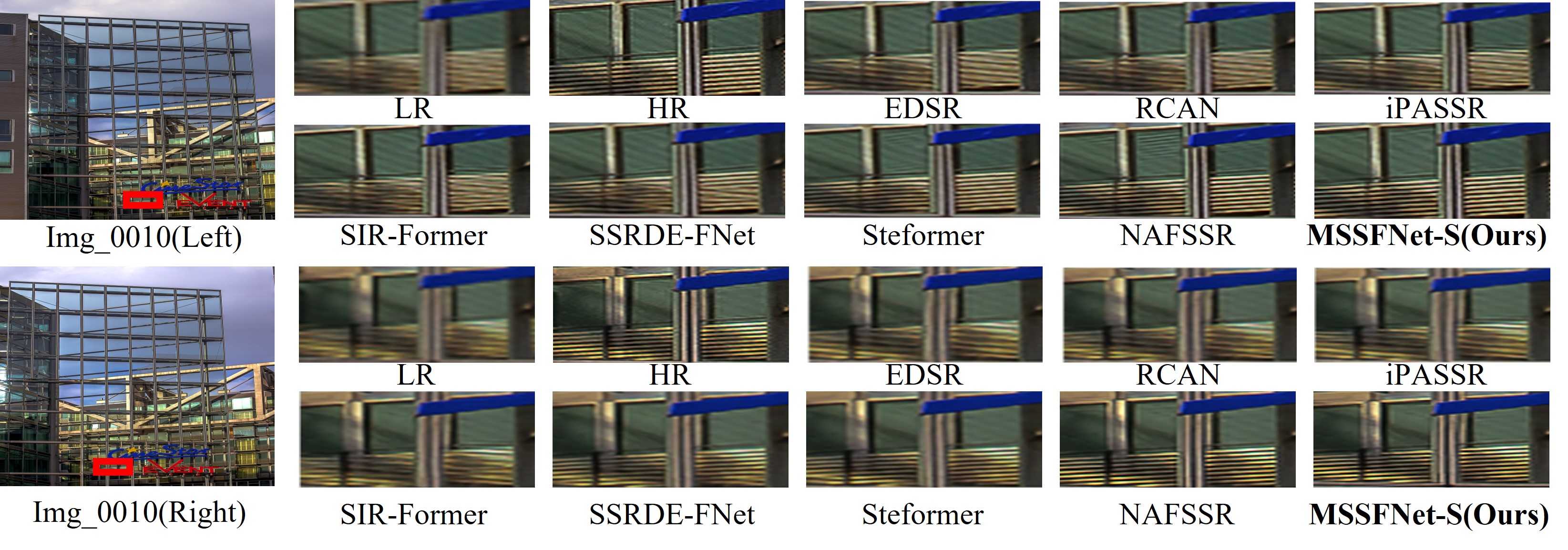}
    \caption{Visual results ($\times 4$) achieved by different methods on the Flickr1024 dataset~\cite{flickr1024wang2019learning}.}
    \label{fig:4x}
\end{figure*}

\begin{figure*}
    \centering
    \includegraphics[width=1\linewidth]{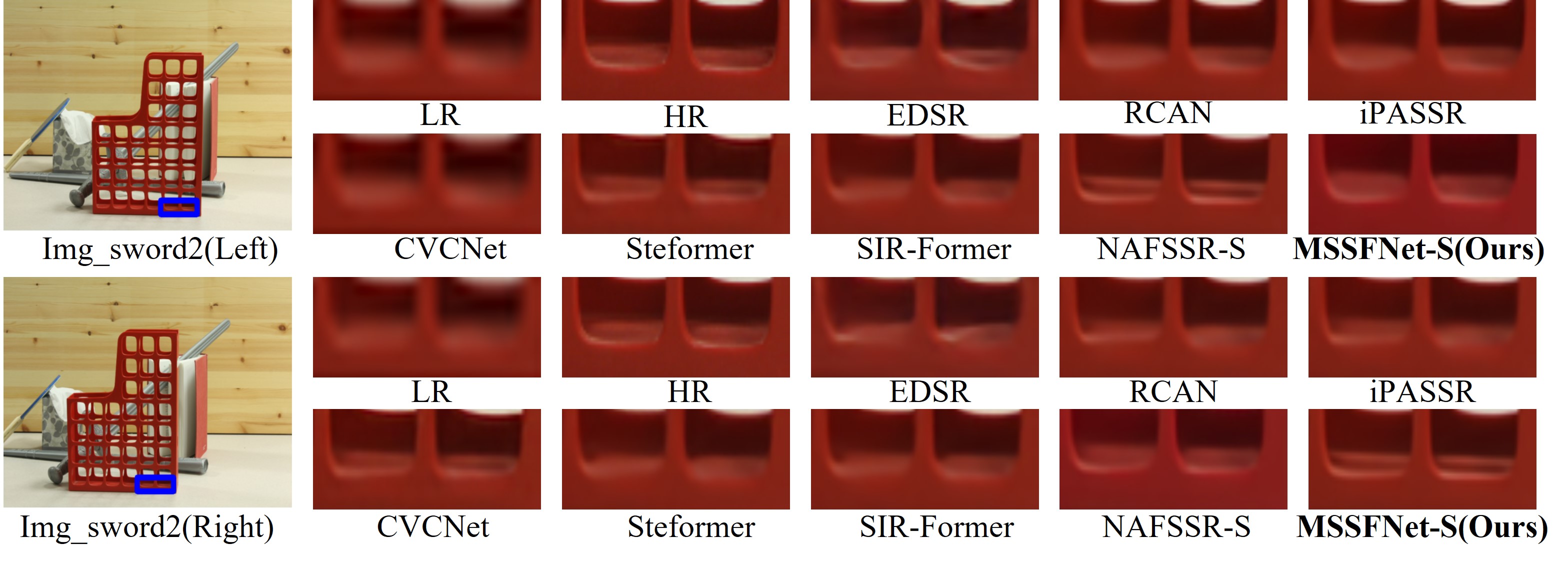}
    \caption{Visual results of different methods for $\times 4$ SR on the Middlebury dataset~\cite{middleburyscharstein2014high}.}
    \label{fig:4xv}
\end{figure*}

\textbf{Visual Comparison.}
We present visual comparisons in Figures~\ref{fig:2xv},~\ref{fig:4x} and~\ref{fig:4xv}. These figures illustrate that our MSSFNet adeptly generates visually pleasing super-resolution images with intricate details and well-defined edges. In contrast, many other methods either produce overly smooth images, sacrificing structural content and fine textural details, or result in images with chroma artifacts and blotchy texture. This solidifies the evidence of the effectiveness of our MSSFNet.

\subsection{Ablation Study}
\begin{figure*}
    \centering
    \includegraphics[width=1\linewidth]{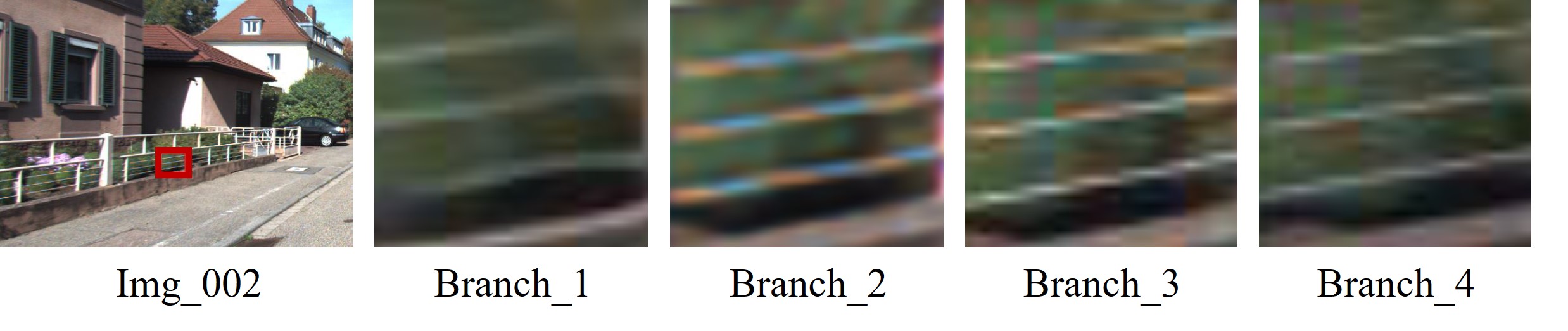}
    \caption{Visual results ($\times 2$) achieved by different experimental designs  incorporating different branch structures on the KITTI 2012 dataset~\cite{kitti2012geiger2012we}. Branch$_i$ represent MSB has $i$ branches.}
    \label{fig:msbabl}
\end{figure*}

\textbf{Mixed-sacle Block.}
To verify MSB's capability in preserving precise spatial details and capturing abundant contextual features, we conducted several experimental designs incorporating different branch structures. As shown in Table.~\ref{tab:msbabl}, compared to a single branch, the utilization of multiple branches in MSB leads to improved performance, indicating the effective extraction of contextual information. As the number of branches increases, the number of model parameters also increases, but the performance does not improve. To further investigate the reasons behind this observation, we present plots of the super-resolution results obtained from different experimental designs (see Figure~\ref{fig:msbabl}). From the Figure~\ref{fig:msbabl}, it is evident that as the number of branches increases, the overall structure of the results remains stable. However, there is a noticeable degradation in the accuracy of capturing fine details.
This observation indicates that when an excessive amount of multi-scale information is captured, the weight assigned to contextual features in the final feature representation becomes disproportionately higher compared to the weight assigned to spatial detail features. As a result, this imbalance in feature weighting can negatively impact the overall performance of the model.

\begin{table}
    \caption{The capability of mixed-scale block. The branch number indicates the number of branches the MSB has, while the kernel size determines the size of the convolution kernel employed in processing each branch. Here we report the results in both PSNR and SSIM for $2 \times$ SR.}
    \label{tab:msbabl}
    \centering
    \resizebox{\linewidth}{!}{
    \begin{tabular}{ccccc}
    \hline
         Branch number&  Kernel size & $\#P$ &  PSNR & SSIM 
         \\
         \hline
        1 & 3x3 & 0.54M & 35.45&  0.9550
         \\ 
         2 & 3x3, 5x5 &0.57M & \textbf{35.82}&\textbf{0.9553}
         \\
         3 & 3x3, 5x5, 7x7&0.98M  & 35.79&0.9551
        \\
        4 & 3x3, 5x5, 7x7, 9x9 &1.06M& 35.77 & 0.9549
         \\
         \hline
    \end{tabular}}
\end{table}

\textbf{Selective Fusion Attention Module.}
To demonstrate the effectiveness of SFAM, we replace the cross-attention module in MSSFNet with several state-of-the-art (SOTA) approaches, including biPAM~\cite{ipassrwang2021symmetric}, SCAM~\cite{nafssrchu2022nafssr}, and RCAM~\cite{swinfsrChen_2023_CVPR}. As presented in Table~\ref{tab:sfamabl}, compared to biPAM, SCAM, and RCAM, our SFAM achieves improvements of 0.23 dB, 0.2 dB, and 0.18 dB, respectively. Additionally, when examining MSSFNet without SFAM to assess the impact of the proposed SFAM on cross-view information, our method achieves a 0.4 dB improvement with SFAM.
\begin{table}
    \caption{The influence of different cross-attention modules. We
here report the results in both PSNR and SSIM for $4 \times$ SR.}
    \label{tab:sfamabl}
    \centering
    \begin{tabular}{cccccc}
    \hline
         Modules&  -&  biPAM & SCAM & RCAM & \textbf{SFAM (Ours) }
         \\
         \hline
         PSNR& 23.41 & 23.58 &23.61  &23.63  &\textbf{23.81}
         \\ 
         \hline
         SSIM& 0.7192 &0.7382  & 0.7409 & 0.7411 &\textbf{0.7418}
         \\
         \hline
    \end{tabular}
\end{table}

\begin{table}
    \caption{The influence of 'selection'. We here report the results in PSNR  for $2\times$ SR. Noted that, "w/" denotes the usage of 'selection', while "w/o" stands for direct calculation of attention.}
    \label{tab:selecabl}
    \centering
    \begin{tabular}{c|c|cc}
    \hline
        \multicolumn{2}{c|}{Methods}& PSNR  & $\triangle$ PSNR 
        \\
        \hline
        \multirow{2}{*}{NAFSSR~\cite{nafssrchu2022nafssr}}
        & w/o & 35.30  & -
        \\
        & w & 35.65  & +0.35
            \\
            \hline
              \multirow{2}{*}{Steformer~\cite{steformer10016671}}
        & w/o & 35.23  & -
        \\
        & w & 35.61  & +0.38
            \\
        \hline
        \multirow{2}{*}{MSSFNet}
        & w/o&35.46  & -
        \\
 
        & w & 35.82  & +0.36
         \\
    \hline
    \end{tabular}
\end{table}

In addition, in order to verify the effect of the "selection" before the attention, ,we conduct experiments on the existing method NAFSSR~\cite{nafssrchu2022nafssr}, Steformer~\cite{steformer10016671}, and our MSSFNet. As shown in Table~\ref{tab:selecabl},  it is evident that the PSNR values for NAFSSR~\cite{nafssrchu2022nafssr}, Steformer~\cite{steformer10016671} and MSSFNet without "selection" are $35.30$, $35.23$ and $35.46$, respectively. 
After adding the selection operation, NAFSSR, Steformer and LSSR achieve a PSNR improvement of $+0.35$ dB, $+0.38$ dB and $+0.36$ dB,  respectively. This suggests that nearby pixels tend to be more similar to each other. The "selection" operator helps reduce irrelevant context from long-range pixel dependencies. By discarding smaller similarity weights from some long-range feature interactions before calculating self-attention, this step enables more accurate representation, leading to higher-quality output. Furthermore, it demonstrates that the "selection" operation  can serve as a general concept applicable to other models, leading to performance improvements.

\begin{table}
    \caption{The influence of fast fourier convolution block, FFCB-1 and FFCB-2 denote FFCB in early and finale stages. We
here report the results in both PSNR and SSIM for $2 \times$ SR.}
    \label{tab:ffcbabl}
    \centering
    \begin{tabular}{cccc}
    \hline
      FFCB-1 &FFCB-2 & PSNR & $\triangle$PSNR
         \\
         \hline
          \faTimes& \faTimes  & 35.09 &- 
          \\
          \faCheck & \faTimes & 35.20 & +0.11
          \\
          \faTimes & \faCheck & 35.16 & +0.07
          \\
          \faCheck & \faCheck & 35.24 & +0.15
         \\
         \hline
    \end{tabular}
\end{table}

\textbf{Fast Fourier Convolution Block.}
To assess the effectiveness of FFCB, we conduct experiments using different model variants, as shown in Table~\ref{tab:ffcbabl}. The results clearly demonstrate that FFCB brings additional performance benefits by capturing global information. Furthermore, we observe that placing FFCBs at different locations in the network pipeline has specific impacts on super-resolution performance. Simultaneously applying FFCB in both early and final stages leads to an improvement in PSNR for MSFFNet, from 35.09 dB to 35.24 dB.

\textbf{Data augmentations.}
We conducted experiments on our model using various data augmentations to assess their effectiveness. The results in Table.~\ref{tab:aug} demonstrate that the inclusion of data augmentations, such as random horizontal flip, random vertical flip, and channel shuffle, positively influences the performance of MSSFNet. When all three data augmentations were applied simultaneously, the PSNR of MSFFNet improved from 23.42 dB to 23.81 dB. This enhancement of 0.08 dB surpasses the improvement achieved by using random flip alone.
\begin{table}
    \caption{Results achieved  with different data augmentations on Flickr 1024~\cite{flickr1024wang2019learning} }
    \label{tab:aug}
    \centering
    \resizebox{\linewidth}{!}{
    \begin{tabular}{ccccc}
    \hline
         Horizontal flip&  Vertical flip&   Channel shuffle&  PSNR& $\triangle$PSNR
         \\
         \hline
         \faTimes& \faTimes &  \faTimes& 23.42 &- 
         \\
         \hline
         \faCheck&  \faTimes&  \faTimes&  23.62& +0.20
         \\
         \faTimes&  \faCheck& \faTimes & 23.62 & +0.20
         \\
         \faTimes& \faTimes &\faCheck  &23.61  &+0.19
         \\
         \hline
         \faCheck&\faCheck  &\faTimes  &23.73  &+0.31
         \\
         \faCheck& \faCheck &\faCheck  &23.81 &+0.39
         \\
         \hline
    \end{tabular}}
\end{table}

\begin{table}
    \caption{Results achieved  with different input schemes for $4\times$ SR on KITTI 2015~\cite{kitti2015menze2015object}. Here, we report the results in both PSNR and SSIM of the cropped left views.}
    \label{tab:inpalbl}
    \centering
    \begin{tabular}{ccc}
    \hline
         Models&  Inputs& PSNR/SSIM
         \\
         \hline
         with single input& Left &25.55/0.7843
         \\ \hline
         with replicated inputs&  Left-Left& 25.62/0.7877
         \\
         \hline
         MSSFNet &  Left-Right& 25.96/0.7946
         \\
         \hline
    \end{tabular}

\end{table}

\textbf{Single Input vs. Stereo Input.}
StereoSR  leverages supplementary data from cross-view images to significantly improve performance  compared with SISR. To showcase the efficacy of stereo information in enhancing super-resolution performance, we conducted experiments using various input schemes. As indicated in Table~\ref{tab:inpalbl}, utilizing individual images during training results in a PSNR decrease of $0.45$ dB compared to the baseline network. Likewise, when employing duplicated left images as inputs, the performance of this modified configuration notably falls short of our initial network. These trials underscore the efficacy of our MSFFNet in capturing information from various perspectives.

\section{Conclusion}
In this paper, we present a mixed-scale selective fusion network (MSSFNet) designed for stereo super-resolution (stereoSR), effectively capturing rich and precise intra-view information while selectively integrating the most accurate cross-view details. Specifically, we introduce the mixed-scale block (MSB) to preserve intricate spatial details while obtaining contextually enhanced feature representations. Additionally, the selective fusion attention module (SFAM) dynamically facilitates the exchange of accurate information between intra-view and cross-view features. To seamlessly integrate local and global information, we introduce the fast Fourier convolution block (FFCB), providing collaborative refinement for the MSB. Furthermore, to reduce computational costs, we replace nonlinear activation functions with  simple gate (SG). Extensive experiments demonstrate the effectiveness of the our MSSFNet.

\bibliographystyle{IEEEtran}
\bibliography{refbib}

\vfill

\end{document}